\documentclass[12pt, letterpaper, onecolumn]{IEEEtran}

\usepackage{times}
\usepackage{epsfig}
\usepackage{graphicx}
\usepackage{subfigure}
\usepackage{amsmath}
\usepackage{amssymb}
\usepackage{url}
\usepackage{pifont}
\usepackage{setspace}
\usepackage{caption}
\usepackage{float}

\begin{document}
%
\title{Design and Implementation of a 3D Undersea Camera System}
%
%
%

\author{Xida~Chen,
        Steve~Sutphen,
        Paul~Macoun,
        and~Yee-Hong~Yang ~\IEEEmembership{}
\thanks{X. Chen, S. Sutphen, and Y-H. Yang are with the Department of Computing Science, University of Alberta, Edmonton, AB, Canada. e-mail: \{xida, ssutphen, herberty\}@ualberta.ca.}
\thanks{P. Macoun is with University of Victoria, Victoria, BC, Canada. email: pmacoun@uvic.ca.}}
\maketitle

\begin{abstract}
In this paper, we present the design and development of an undersea camera system. The goal of our system is to provide a 3D model of the undersea habitat in a long-term continuous manner. The most important feature of our system is the use of multiple cameras and multiple projectors, which is able to provide accurate 3D models with an accuracy of a millimeter. By introducing projectors in our system, we can use many different structured light methods for different tasks. There are two main advantages comparing our system with using ROVs or AUVs. First, our system can provide continuous monitoring of the undersea habitat. Second, our system has a low hardware cost. Comparing to existing deployed camera systems, the advantage of our system is that it can provide accurate 3D models and provides opportunities for future development of innovative algorithms for undersea research.
\end{abstract}


%
\IEEEpeerreviewmaketitle

\section{Introduction}

\IEEEPARstart{T}{he} study of underwater species and their habitat has been a popular research topic in many areas of science and engineering for many decades. Researchers from different disciplines study them for different purposes. For example, marine biologists try to understand the marine environment which helps to support life on earth \cite{Sanmartin2008}. Studying underwater species helps them to understand the effects of climate change, pollution, and invasive species.

A typical technique for undersea monitoring uses either an Autonomous Underwater Vehicle (AUV) \cite{Blidberg2001} or a Remotely Operated Vehicle (ROV) \cite{Martin2005} to explore the underwater habitat. There are several disadvantages of using this technique. First, AUVs or ROVs cannot be permanently deployed undersea. Instead, it usually goes underwater once every 4 to 6 months. Therefore, the collected data can only be compared with that collected several months ago, which may not be useful in some studies. Second, temporal data cannot be obtained for the same location because of the lack of landmarks undersea and the accuracy of GPS. Third, the cost of using AUVs or ROVs is very high. In particular, a large ship is usually required to carry the AUVs or ROVs out to the sites. The entire crew of the ship is also required, and several biologists are typically onsite to collect data. Finally, the expedition is at the mercy of the weather.

In this paper, we describe a system that is deployed permanently undersea. The goal of this system is to perform long-term monitoring of the underwater habitat. Comparing to using AUVs or ROVs, the cost of our system is significantly lower. Moreover, our system is able to provide information of the underwater habitat on demand or on an hourly or daily basis. Additionally, we have the option of streaming data in real time. There are three challenges of developing such a system.

\ding{182} We need to provide accurate 3D models of the undersea habitat from 2D images. In order to achieve high accuracy 3D reconstruction, an appropriate method has to be selected.

\ding{183} Because of the above requirement, the hardware selection is another major challenge.

\ding{184} Since our system monitors the undersea habitat regularly, the amount of data is huge and the storage of it is another challenge.

\section{Literature Review}
Using ROVs or AUVs \cite{Greene1993, Griffiths2002} to explore the ocean is still a popular approach nowadays. They are useful in many ways, mostly because of human safety and the ability to conduct continuous operations. In the intense underwater pressure of the marine environment, ROVs or AUVs eliminate the risk of accidents to the occupants of manned submersibles. The failure of pressure seals or of the life-support system that provides oxygen and removes carbon dioxide is not a concern in ROVs or AUVs. Moreover, ROVs can operate in a continuous manner. In contrast, due to safety concerns, shipboard recovery of these submersibles can often restrict operations to daylight hours. Despite of their advantages, there are still limitations of using ROVs or AUVs to explore the ocean. First, ROVs or AUVs are usually deployed undersea every 4 to 6 months. During each trip, the digital camera(s) and video camera(s) mounted on them are used to collect as much data as possible. However, one fundamental issue is that the data is only available every 4 to 6 months and there is no data in between. Furthermore, it is difficult if not impossible to go back to exactly the same location because of the lack of landmarks and the accuracy of the GPS. In this case, the data may not be very helpful to biologists or ecologists. For example, in order to study how the pollution impact the undersea habitat, the image or video data may need to be continuous. Furthermore, the cost of each trip is very high. In particular, in addition to the time for planning, a typical ROV trip takes 3-7 days to explore different observation sites. A large ship and its entire crew are needed to carry the ROV to the observation site. Additionally, a group of scientists has to be onsite to navigate the ROV in order to collect data at the location that would be useful to them. Last but not least, the trip can also be impacted by weather. When the undersea turbulence is too strong due to bad weather, the ROV will be very hard to control. Careless control of the ROVs or AUVs could damage both the vehicle and the undersea habitat.

Due to the above limitations of ROVs and AUVs, there are systems being developed to long-term surveillance purpose. A camera surveillance system is presented in \cite{Boom2012} to monitor the fish populations. It focuses on the development of software to recognize and track fish from a very large database of videos. Ocean Networks Canada \cite{Heesemann2014, Barnes2013} has deployed many undersea systems along the western coast of Canada. Some systems are deployed at over 2000m deep, while others are at about 100m deep. The entire Ocean Networks Canada consists of NEPTUNE and VENUS cabled observatories, where NEPTUNE is an earthquake and tsunami research lab and VENUS an underwater landslide research lab. The collected data are shared with many researchers for data visualization and analysis. The data is potentially useful for many applications, such as studying ocean/climate change, ocean acidification, recognizing and mitigating natural hazards, non-renewable and renewable natural resources. With such a huge observation network, different functionalities can be achieved. Consider NEPTUNE as an example. It can be used to determine the presence of fish sounds by examining the deep sea acoustic recordings \cite{Wall2014}. It is also used to study faunal grouping \cite{Juniper2013}. The study shows that distinct seasonal faunal groupings are observed, together with summer and winter trends in temperature, salinity and current patterns. We can see that a long-term surveillance system can achieve much more than that from discrete ROV or AUV trips.

Motivated by the NEPTUNE and VENUS systems, we design and implement a camera system that is currently deployed undersea to monitor the habitat. There are several unique features of our system. First, our system operates in a continuous manner, captures images on demand or on predefined intervals and provides 3D model of the habitat. Second, our system is designed to be easily extended for future development. Last but not least, the cost of our system is low. The total hardware cost is less than \$140,000 (CAD).

\section{System Overview}
The goal of our system is to provide a 3D model of the undersea habitat from 2D images. To achieve that, our system needs an image capture module and a control module. The image capture module is in charge of taking images from different views while the control module is in charge of streaming and collecting data from each camera.

\subsection{Image Capture Module}
The most important feature of our system is the introduction of multiple projectors and multiple cameras. We apply a structured light method to build the 3D model from 2D images. A structured light method is applied because of its high accuracy, which can typically limit the error to within a millimeter. The method requires at least one projector and one camera. We use multiple camera and multiple projectors so that the accuracy is even higher, and the observed area is larger. Using projectors in our system is a critical design for future extension. That is, projectors can be used to project different structured light patterns for accurate 3D object reconstruction. There are many structured light methods that can be used in our system \cite{Geng2011}. Some of the methods can be used to capture dynamic events, while others can achieve extremely high accuracy. In other words, our system is a framework that can be used to serve multiple purposes.

\begin{figure}[h!]
\centering
   \includegraphics[width=1\linewidth]{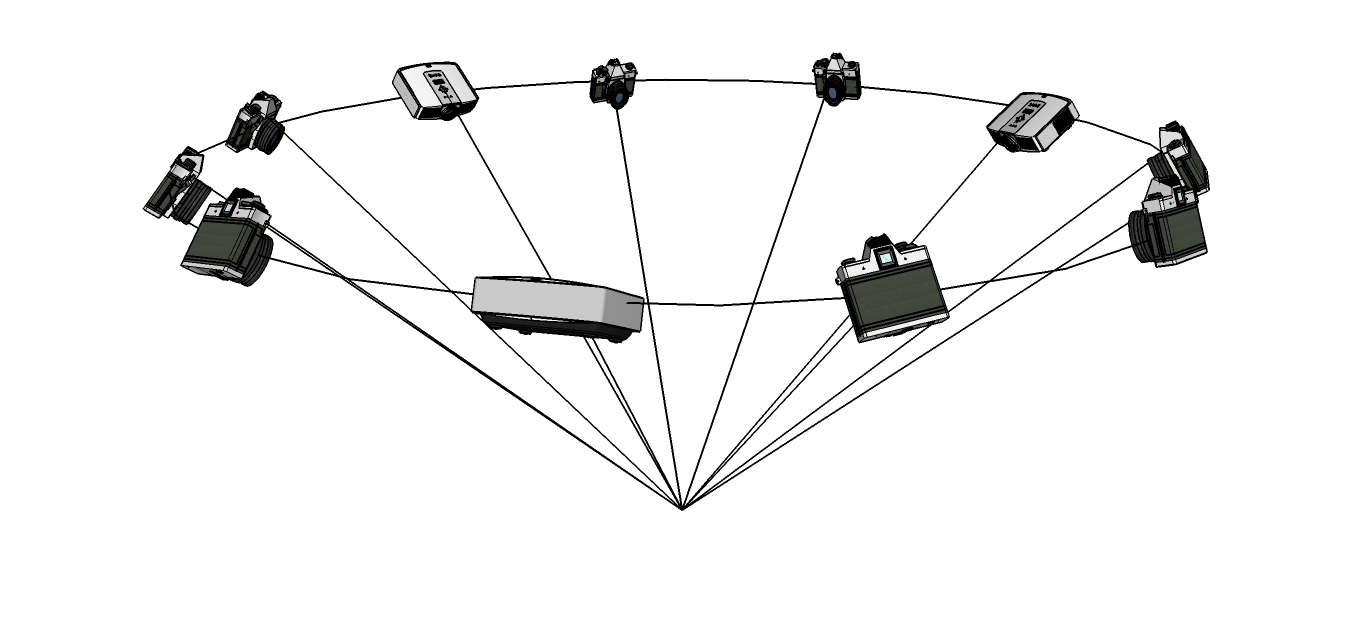}
   \caption{Diagram of the image capture module.}
\label{fig:imageCaptureDiagram}
\end{figure}
Fig. \ref{fig:imageCaptureDiagram} shows the diagram of our image capture module. There are 8 cameras and 3 projectors arranged in a full circle, and all the devices are looking at a common point. The devices are arranged so that the system can observe as much information of the habitat as possible. Each camera used in our system is a Point Grey Flea3 GigE color camera with 5M pixels, where the data can be streamed through the Ethernet. This camera is selected because it can be easily controlled through the Ethernet and the image buffer can be transferred quickly. Since each device is placed in a watertight housing, ideally the device should generate no or very low heat. It is the main reason that we select the TI Lightcrafter as our projector, which is a pico data projector that can operate without a fan. Another reason is that TI Lightcrafter can be controlled via the Ethernet for projecting structured light patterns, which is consistent with the camera. It can potentially achieve a very high frame rate at over 1000 fps.

\subsection{Control Module}
\begin{figure}[htb]
\centering
   \includegraphics[width=1\linewidth]{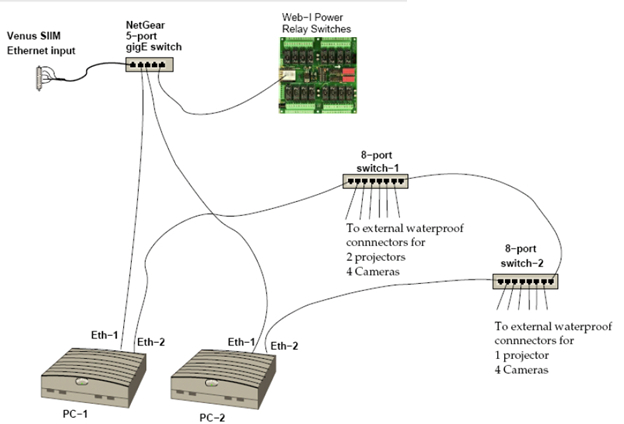}
   \caption{Diagram of the control module.}
\label{fig:controlDiagram}
\end{figure}
Fig. \ref{fig:controlDiagram} shows the diagram of our control module. There are two PCs for redundancy in charge of controlling the image capture module, where one of them is a main operating PC and the other one serves as a backup in case of hardware failure. In particular, the PC is running the software that controls the TI Lightcrafter to project structured light patterns and the cameras to capture images simultaneously, as well as collecting buffered data from the cameras. In our design, the PC is carefully selected to satisfy several requirements. First of all, the PC needs to communicate with the image capture module through the Ethernet. It also needs to communicate with the server in order to upload the data. As well, the PC should be fast. In this case, we select a PC that has at least two Ethernet ports, one for each of the above functions so that the two separate communication channels will not interfere with each other. Another important requirement of the PC is that it must run without a fan. The cooling for the hardware such as the CPU should be air cooled and very efficient. As a result, the Intense PC \cite{IntensePC} is selected because it fits all of our requirements. It is a fanless PC with an Intel i7 CPU, 8GB memory, two 1Gb Ethernet ports, and 500GB of solid state hard drive. In the diagram, we can see that one of the Ethernet ports of the PC (Eth-2) is connected to two 8-port Ethernet switch, which connects 8 cameras and 3 projectors together. Another Ethernet port (Eth-1) is connected to a 5-port switch which connects to an on-shore computer as well as to a Web-I power relay board [12]. Using the Web-I relay board is another unique design of our system. In particular, the board is software programmable and has 16 ports acting as power switches for the 8 cameras, 3 projectors, 2 PCs and 2 8-port switches. With this design, we can use the on-shore computer to turn on/off any device in our system through a program. Moreover, once one of the PCs is turned on, we can also use programs to turn on/off any device from the undersea PC, which gives more control on the image capture module. For example, we may not want the projector to be turned on at all time because the light will attract fish.

\subsection{Housing}
\begin{figure}[htb]
\centering
   \includegraphics[width=1\linewidth]{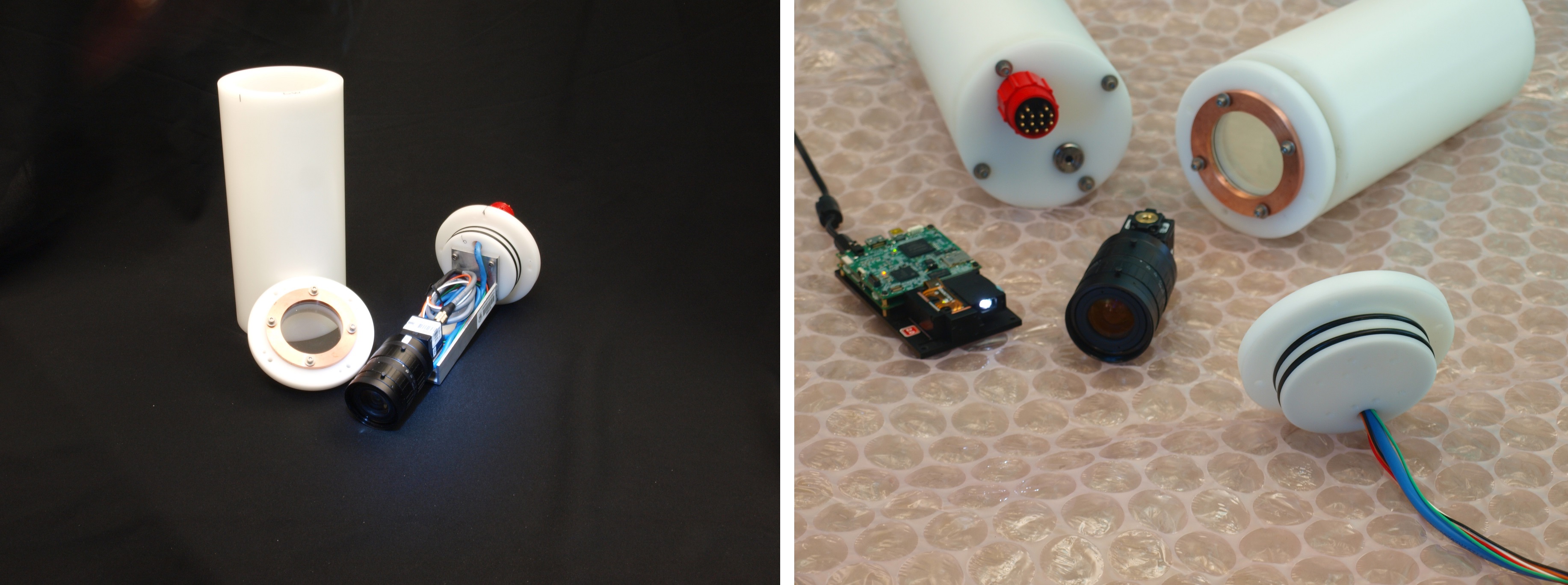}
   \caption{Left: A camera and its housing. Right: A camera, a projector and their housings.}
\label{fig:housing}
\end{figure}

Each camera or projector is mounted in a watertight housing equipment. Moreover, the entire control module is sealed in a watertight housing equipment. The left image in Fig. \ref{fig:housing} shows a camera and its housing before the camera is sealed inside. The housing equipment is tested to guarantee that it can resist a pressure of 100m depth of water. The right image shows a camera, a projector and their housings.

\begin{figure}[htb]
\centering
   \includegraphics[width=1\linewidth]{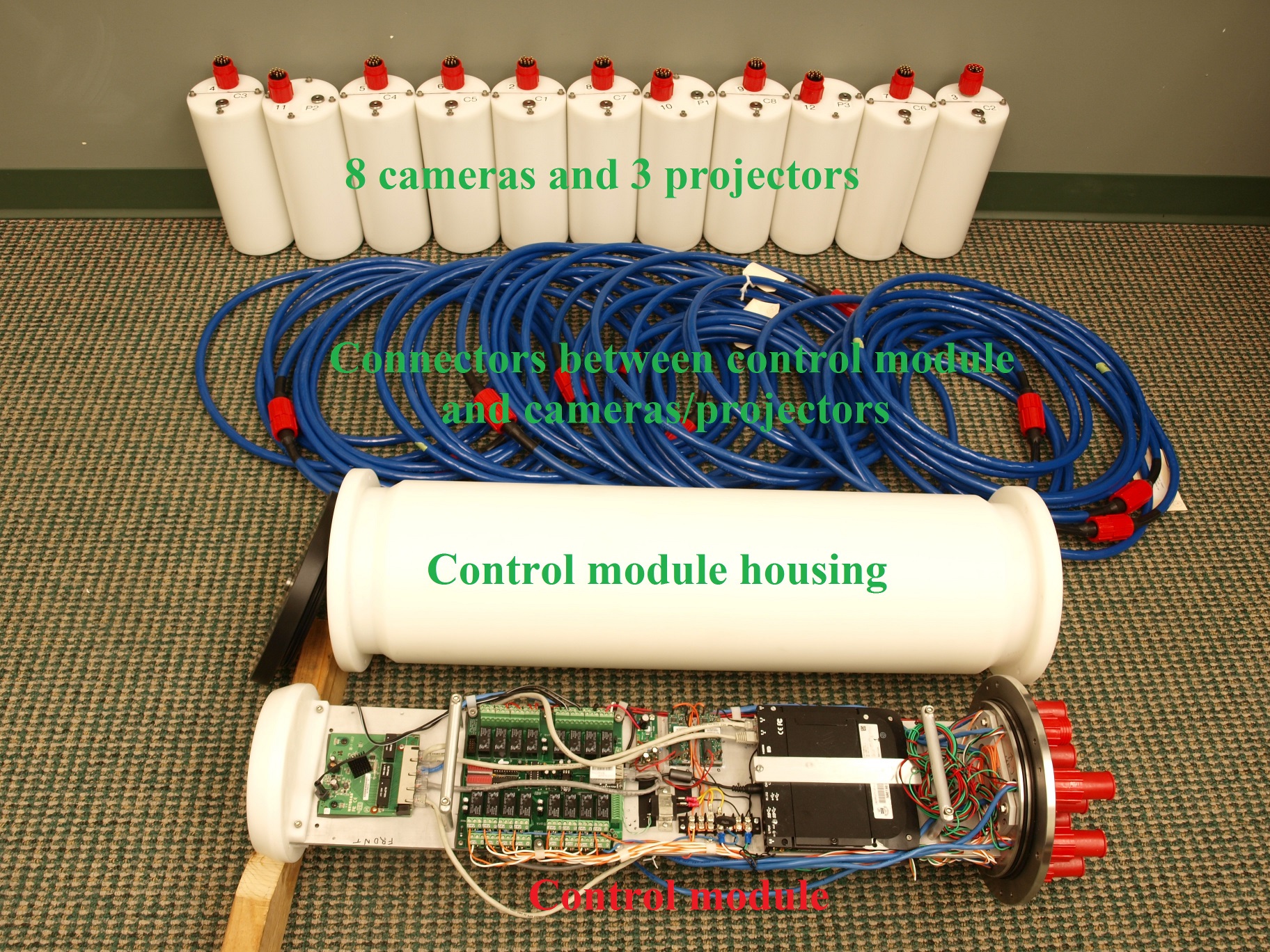}
   \caption{The system before its final assembly.}
\label{fig:systemOnBench}
\end{figure}
\begin{figure}[htb]
\centering
   \includegraphics[width=1\linewidth]{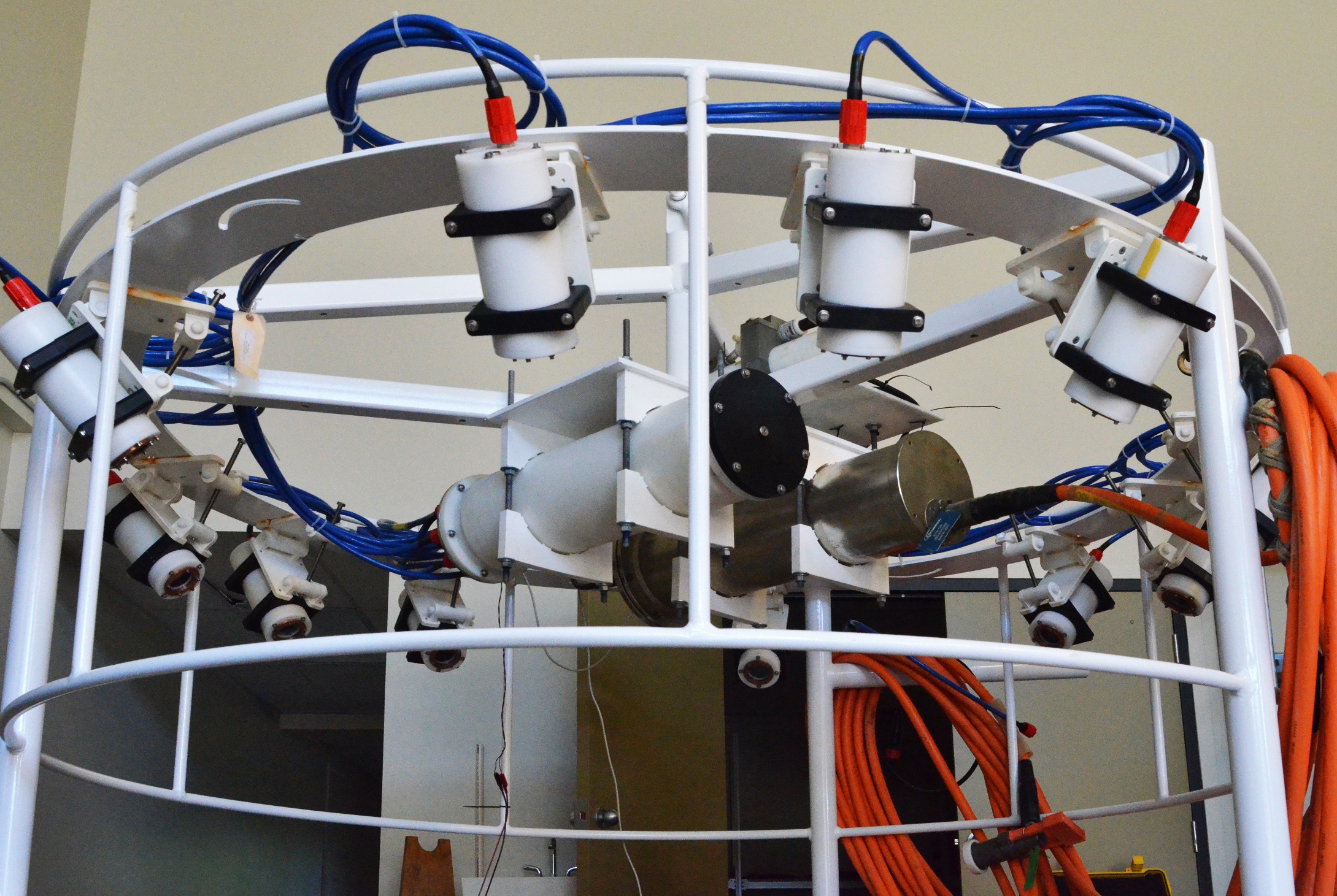}
   \caption{The system before it is deployed undersea.}
\label{fig:systemBeforeDeploy}
\end{figure}

Fig. \ref{fig:systemOnBench} shows the system before assembling. In particular, the 8 cameras and 3 projectors are put inside the housing and standing next to the connectors. The control module and its housing are also shown in the figure. In this figure, we can only see one side of the control module. On this side, there is one PC, the relay board and the 5-port switch attached. On the other side, the other PC, the two 8-port switches are attached. With this design, the space usage is minimized. Fig. \ref{fig:systemBeforeDeploy} shows all the components of our system after they have been mounted on a supporting frame and is ready to be deployed undersea.

\section{Software Development}
The software of our system includes two parts: system calibration and 3D reconstruction.

\subsection{System Calibration}
In order to obtain accurate 3D models for the underwater habitat, accurate system parameters are needed which are obtained by system calibration. In particular, there are two sets of parameters that are required for 3D reconstruction. The first includes the camera intrinsic parameters and the relative pose of cameras \cite{Hartley2004}. The second consists of the refractive interface normal and the distance from the camera center to the refractive interface for each camera \cite{Chen2014}. Both sets of parameters can be obtained when we built the system and before it is deployed undersea. The second set is used to accommodate for water refraction so that the 3D reconstruction result is more accurate.

\subsection{3D Reconstruction}
We use the structured light method for 3D reconstruction of the undersea habitat, to be more specific, the gray code method \cite{Geng2011}. After 3D reconstruction, the captured image as well as the 3D model is uploaded to a data server instead of storing in the PC.

\section{Results}
\begin{figure}[htb]
\centering
   \includegraphics[width=1\linewidth]{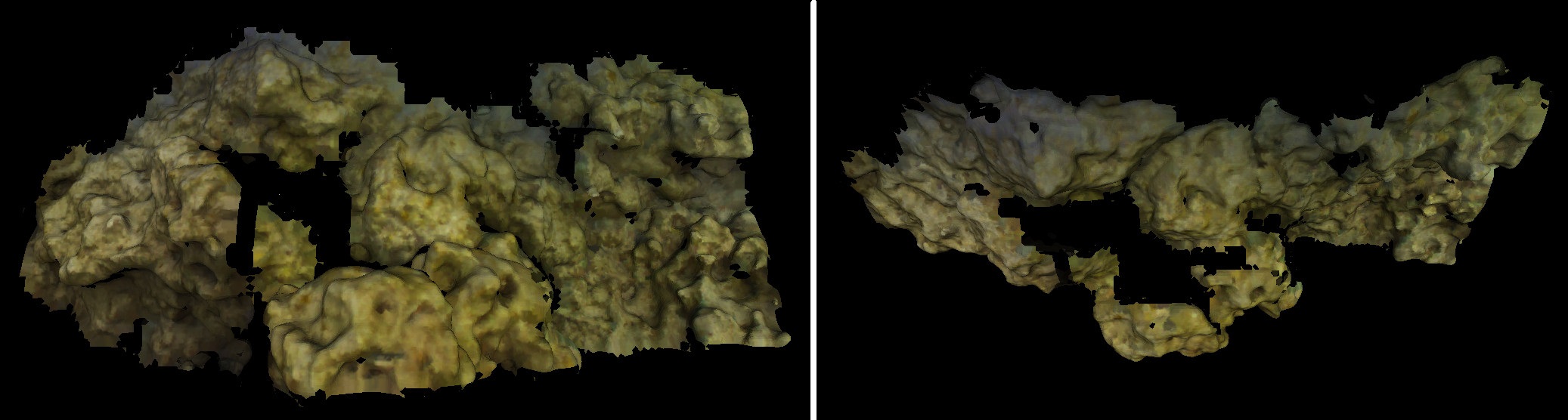}
   \caption{3D reconstruction results of a coral reef in the lab: front view (left) and top view (right).}
\label{fig:ourResultLab}
\end{figure}

\begin{figure}[htb]
\centering
   \includegraphics[width=1\linewidth]{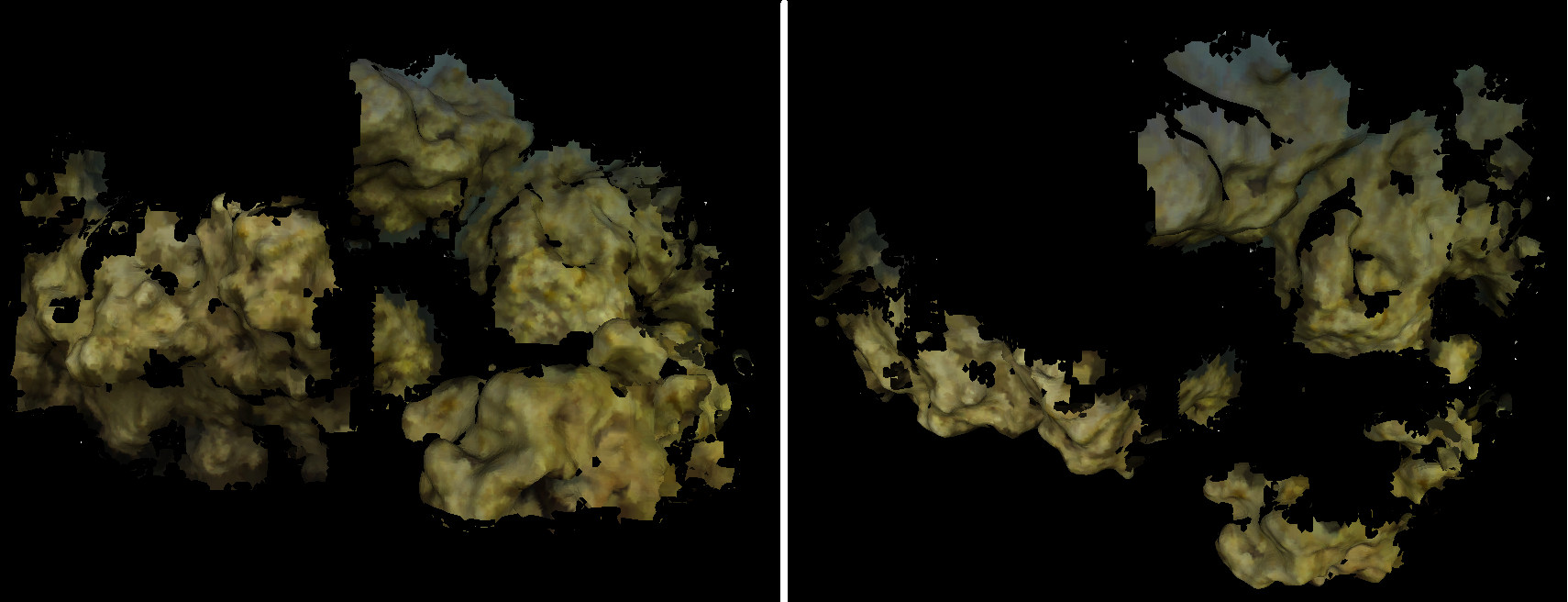}
   \caption{3D reconstruction results using PMVS2 \cite{Furukawa2010}. Left: front view. Right: Top view.}
\label{fig:PMVS2ResultLab}
\end{figure}

The prototype of our system was tested in our lab. In particular, Fig. \ref{fig:ourResultLab} shows 3D reconstruction result of a coral reef placed in water by using the prototype. The left image shows the result from the front view and right image from the top view. We can see that a lot of the surface details are captured in the result by using our structured light method. Fig. \ref{fig:PMVS2ResultLab} shows result without using structured light patterns \cite{Furukawa2010}. We can see that our method is able to provide much better result than that of \cite{Furukawa2010}.

\begin{figure}[htb]
\centering
   \includegraphics[width=1\linewidth]{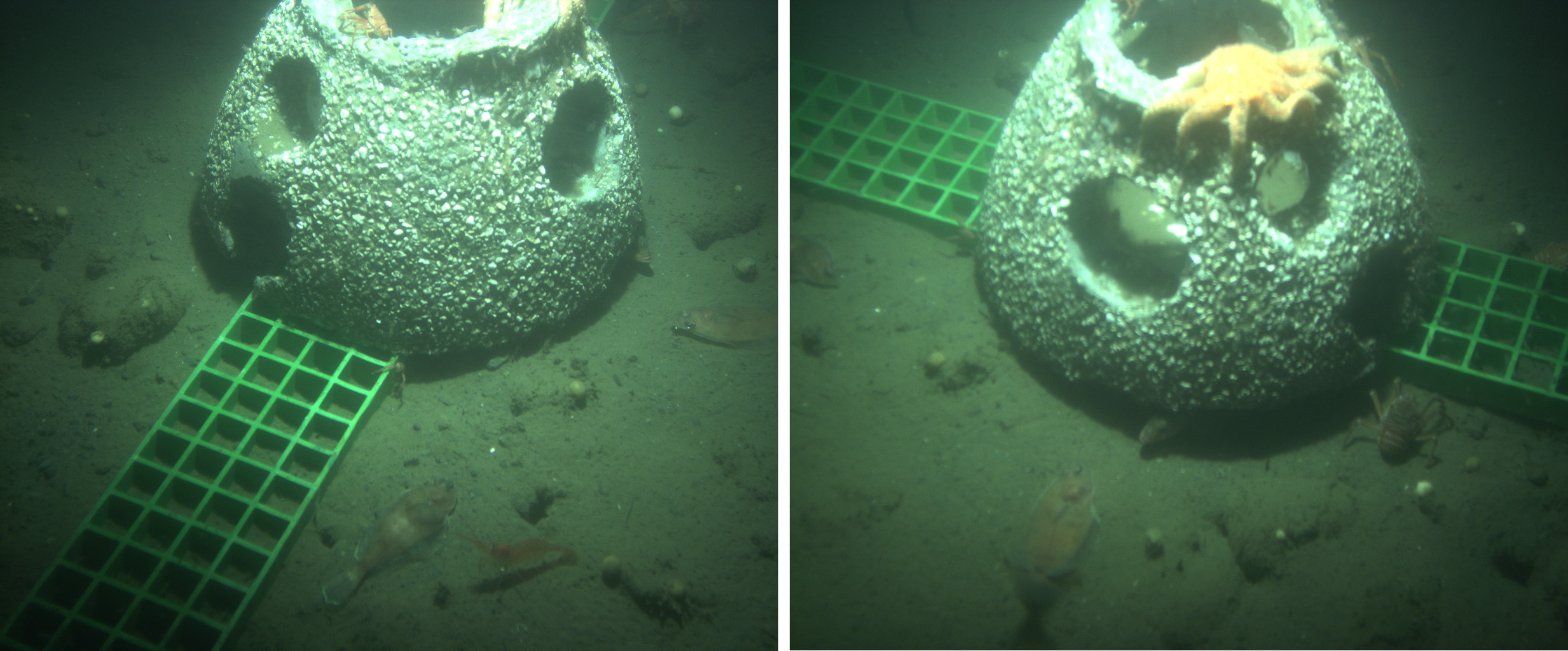}
   \caption{Images captured by two different cameras.}
\label{fig:waterImgs}
\end{figure}

\begin{figure}[htb]
\centering
   \includegraphics[width=1\linewidth]{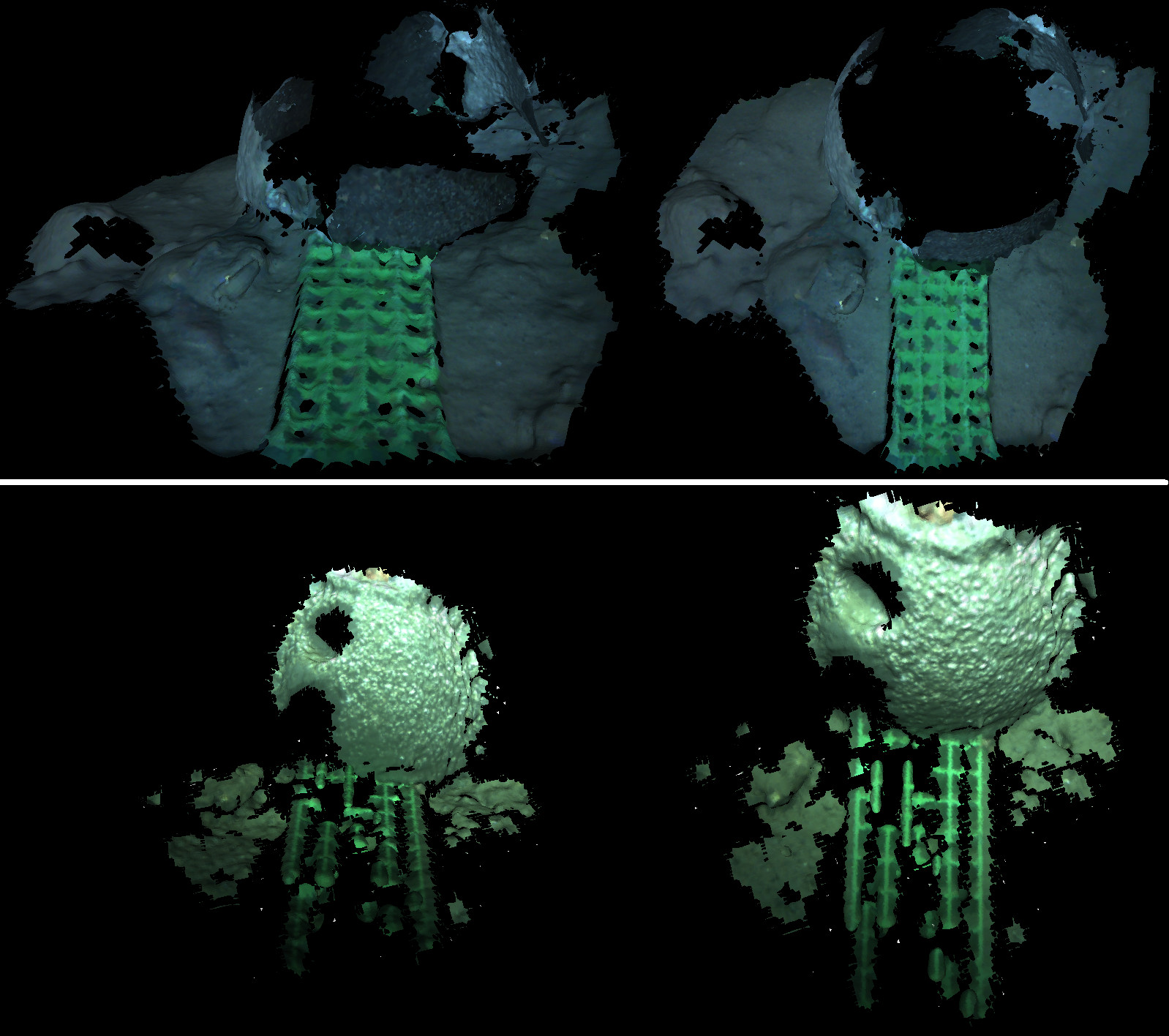}
   \caption{Top row: 3D reconstruction result by our method. Button row: result by using the method in \cite{Furukawa2010}.}
\label{fig:realRes}
\end{figure}

Our system starts capturing images once it is deployed undersea. Fig. \ref{fig:waterImgs} shows images captured by two different cameras. Moreover, Fig. \ref{fig:realRes} compares the 3D reconstruction results of using our method with that of using the method presented in \cite{Furukawa2010}.

\section{Conclusion and Future Work}
We design and develop a camera system for long-term monitoring of underwater habitat. We have demonstrated the unique and important feature of our system. Future work of our system includes using different patterns to complete different tasks. In our 3D reconstruction result (top row of Fig. \ref{fig:realRes}), the top half of the centered object is missing because our projector is incorrectly oriented by a small amount. This will be corrected in the future for a full 3D reconstruction.


%

\begin{IEEEbiographynophoto}{Xida Chen}
is currently a PhD student in the Computing Science Department at the University of Alberta. His research interests include computer vision and image processing. In particular, he is working on structured light and stereo methods, and their related applications.
\end{IEEEbiographynophoto}

\begin{IEEEbiographynophoto}{Steve Sutphen}
is a Faculty Service Officer in the Computing Science Department at the University of Alberta. His responsibilities are to 1) provide long-term planning, and organizational support for Departmental research, 2) explore and develop new concepts in hardware and software for research support, 3) install and supervise the maintenance of Departmental equipment.
\end{IEEEbiographynophoto}

\begin{IEEEbiographynophoto}{Paul Macoun}
is managing the latest series of projects funded through the CFI Leading Edge Fund. These include Coastal Radar stations in the Strait of Georgia, instruments on BC Ferries, a Profiling station in Saanich Inlet, Gliders, and a host of new bottom mounted sensor systems. Paul brings a varied background in engineering and oceanography to the project. He has field experience in marine seismics and oceanography, and several years of industrial engineering experience. Paul has a B.A.Sc. Mechanical Engineering from the University of Waterloo (1995), and a M.Sc. Physical Oceanography (2003) from the University of Victoria.
\end{IEEEbiographynophoto}

\begin{IEEEbiographynophoto}{Yee-Hong Yang }
received his BSc (first honors) from the University of Hong Kong, his MSc from Simon Fraser University, and his M.S.E.E. and PhD from the University of Pittsburgh. He was a faculty member in the Department of Computer Science at the University of Saskatchewan from 1983 to 2001 and served as Graduate Chair from 1999 to 2001. While there, in addition to department level committees, he also served on many college and university level committees. Since July 2001, he has been a Professor in the Department of Computing Science at the University of Alberta. He served as Associate Chair (Graduate Studies) in the same department from 2003 to 2005.
His research interests cover a wide range of topics from computer graphics to computer vision, which include physically based animation of Newtonian and non-Newtonian fluids, texture analysis and synthesis, human body motion analysis and synthesis, computational photography, stereo and multiple view computer vision, and underwater imaging. He has published over 100 papers in international journals and conference proceedings in the areas of computer vision and graphics. He is a Senior Member of the IEEE and serves on the Editorial Board of the journal Pattern Recognition. In addition to serving as a reviewer to numerous international journals, conferences, and funding agencies, he has served on the program committees of many national and international conferences. In 2007, he was invited to serve on the expert review panel to evaluate computer science research in Finland.
\end{IEEEbiographynophoto}




\end{document}